\documentclass{article}

\usepackage{amsmath,amssymb,amsthm}
\usepackage{sidecap}

\usepackage{amssymb}
\usepackage{amsmath}
\usepackage{amsthm}
\usepackage{amsfonts}
\usepackage{stmaryrd}
\usepackage{graphicx}

\usepackage[
backend=biber,
style=numeric,
uniquelist=false,
maxcitenames=2,
doi=false,
isbn=false,
url=false,
eprint=false,
maxbibnames=99,
sorting=none
]{biblatex}
\AtEveryBibitem{\clearfield{month}}
\AtEveryCitekey{\clearfield{month}}
\AtEveryBibitem{\clearlist{language}}
\renewbibmacro{in:}{}
\usepackage[table,usenames,dvipsnames]{xcolor}
\usepackage[colorlinks,linktoc=all]{hyperref}
\usepackage[all]{hypcap}
\hypersetup{citecolor=black}
\hypersetup{linkcolor=black}
\hypersetup{urlcolor=black}
\usepackage{cleveref}

\usepackage{amsthm}

\DeclareRobustCommand{\mb}[1]{\boldsymbol{#1}}

\DeclareMathOperator*{\argmax}{argmax}
\DeclareMathOperator*{\argmin}{argmin}
\DeclareMathOperator*{\Tr}{Tr}

\newcommand{\mba}{\mb{a}}

\newcommand{\mbv}{\mb{v}}

\newcommand{\mbx}{\mb{x}}
\newcommand{\mby}{\mb{y}}

\newcommand{\mbA}{\mb{A}}
\newcommand{\mbB}{\mb{B}}
\newcommand{\mbC}{\mb{C}}

\newcommand{\mbI}{\mb{I}}

\newcommand{\mbM}{\mb{M}}

\newcommand{\mbP}{\mb{P}}
\newcommand{\mbQ}{\mb{Q}}

\newcommand{\mbS}{\mb{S}}

\newcommand{\mbU}{\mb{U}}
\newcommand{\mbV}{\mb{V}}

\newcommand{\mbX}{\mb{X}}
\newcommand{\mbY}{\mb{Y}}
\newcommand{\mbZ}{\mb{Z}}

\newcommand{\cB}{\mathcal{B}}

\newcommand{\cF}{\mathcal{F}}

\newcommand{\cO}{\mathcal{O}}
\newcommand{\cP}{\mathcal{P}}

\newcommand{\cR}{\mathcal{R}}

\newcommand{\cT}{\mathcal{T}}

\newcommand{\reals}{\mathbb{R}}

\usepackage[preprint,nonatbib]{neurips_2023}

\usepackage[utf8]{inputenc} 
\usepackage[T1]{fontenc}    
\usepackage{hyperref}       
\usepackage{url}            
\usepackage{booktabs}       
\usepackage{amsfonts}       
\usepackage{nicefrac}       
\usepackage{microtype}      
\usepackage{xcolor}         
\usepackage{soul}
\usepackage{appendix}

\title{Soft Matching Distance: A metric on neural representations that captures single-neuron tuning}
\vspace{-8mm}
\author{%
  Meenakshi Khosla \\
  McGovern Institute for Brain Research\\
  Massachusetts Institute for Technology\\
  \texttt{mkhosla@mit.edu} \\
  \And
  Alex H. Williams \\
  Center for Neural Science, New York University \\
  Center for Computational Neuroscience \\ Flatiron Institute \\
  \texttt{alex.h.williams@nyu.edu} 
}

\begin{document}

\maketitle
\vspace{-8mm}
\begin{abstract}
Common measures of neural representational (dis)similarity are designed to be insensitive to rotations and reflections of the neural activation space.
Motivated by the premise that the tuning of individual units may be important, there has been recent interest in developing stricter notions of representational (dis)similarity that require neurons to be individually matched across networks.
When two networks have the same size (i.e. same number of neurons), a distance metric can be formulated by optimizing over neuron index permutations to maximize tuning curve alignment.
However, it is not clear how to generalize this metric to measure distances between networks with different sizes.
Here, we leverage a connection to optimal transport theory to derive a natural generalization based on ``soft'' permutations.
The resulting metric is symmetric, satisfies the triangle inequality, and can be interpreted as a Wasserstein distance between two empirical distributions.
Further, our proposed metric avoids counter-intuitive outcomes suffered by alternative approaches, and captures complementary geometric insights into neural representations that are entirely missed by rotation-invariant metrics.
\end{abstract}

\section{Introduction}

Neural representations of stimuli and actions are often described in terms of ``tuning curves'' of individual neurons.
The most classic example is work by Hubel and Wiesel~\cite{hubel1959receptive,hubel1962receptive}, which found that neurons in the primary visual cortex (V1) of cats were selectively responsive---i.e. ``tuned''---to edges with particular orientations.
However, the utility of tuning curves is less certain in higher-order brain regions involved in navigation, decision-making, and complex sensory processing.
In such areas, neurons often exhibit complex ``mixed selectivity'' to multiple sensory or task features~\cite{Hung2005,hardcastle2017multiplexed,fusi2016neurons}.
Neuroscientists recurrently debate whether tuning curves are meaningful in these situations, and similar debates have recently arisen in the interpretable artificial intelligence community~\cite{Morcos2018,Pospisil2018,cammarata2020curve}.

\begin{figure}
\includegraphics[width=\linewidth]{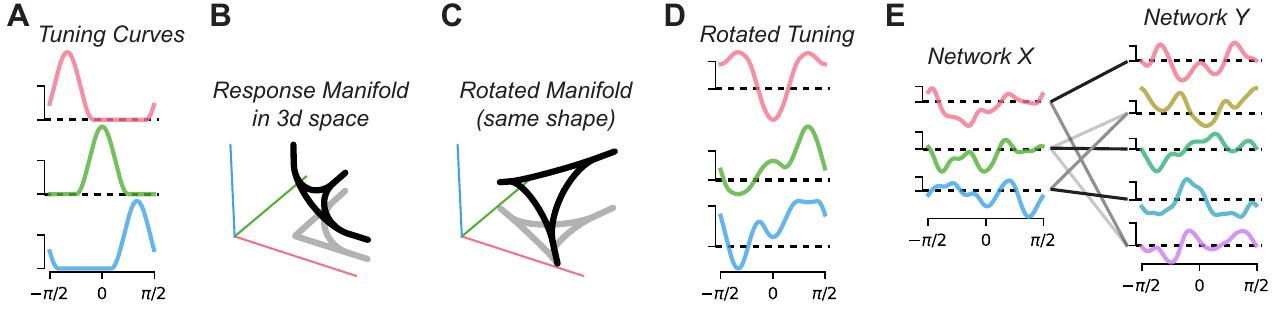} 
\centering
\caption{
    \textbf{(A)} Example tuning curves from 3 neurons over a 1D stimulus space.
    \textbf{(B)} Manifold (black curve) arising from tuning curves from panel A in 3D neural firing rate space. Each coordinate axis encodes a single-neural firing rate, colored as in panel A.
    \textbf{(C)} A rotated manifold with the same shape as panel B.
    \textbf{(D)} Tuning curves associated with the rotated manifold in panel C (compare with panel A).
    \textbf{(E)} Schematic illustration of ``soft matching.'' Grayscale lines show matched similar tuning curves across two networks. The darkness of the line indicates the strength of the match.
}
\label{fig:tuning-schematic}
\end{figure}

The tuning of individual neurons is closely connected to studies of neural geometry~\cite{Kriegeskorte2021-mw}.
In particular, neural tuning curves determine the geometry of population-level neural representations, but the same geometry can be produced by many different sets of tuning curves (Fig.~\ref{fig:tuning-schematic}A-D).
Despite this connection, most investigations implicitly ignore tuning by considering rotation-invariant quantities.
Indeed, popular measures of representational (dis)similarity between neural networks---centered kernel alignment (CKA; \cite{Kornblith2019}), canonical correlations analysis (CCA; \cite{Raghu2017}), representational similarity analysis (RSA; \cite{Kriegeskorte2008}), and Procrustes shape distance~\cite{Williams2021}---are all invariant to rotations of the neural activation space (such as Fig.~\ref{fig:tuning-schematic}B vs. 1C).
Thus, to study the importance of individual neural tuning (or lack thereof) we require complementary metrics that are sensitive to rotations, but still invariant to permutations of the neuron indices (since such indices are often arbitrary).
While a handful of studies have already explored measures that fit these requirements~\cite{Li2015_convergent,Williams2021}, our understanding of these methods and their relation to the more popular approaches cited above is under-developed.

\vspace{-2mm}
For example, \textcite{Williams2021} proposed a rotation-sensitive, permutation-invariant metric on neural representations based on ``permutation Procrustes'' analysis~\cite{gower2004procrustes}.
However, their approach suffers a serious limitation---it can only be applied to pairs of networks with the same number of neurons since it relies on a strict one-to-one matching between units.
The central contribution of our work is to generalize their metric to the much more typical case of unequal network sizes, which we achieve by leveraging elementary principles from optimal transport theory~\cite{Peyre2019} to obtain a \textit{soft matching} or \textit{soft assignment}~\cite{rangarajan1997softassign} between two sets of tuning curves (Fig.~\ref{fig:tuning-schematic}E).
Similar approaches have been used for representation transfer in deep networks \cite{singh2020model, li2020representation} and to align word embeddings \cite{grave2019unsupervised,Alvarez2019}.
Our central motivation---to quantify the reproducibility of tuning curves across networks---is distinct from these prior works, and the details of our approach are suitably adapted where necessary.

Importantly, our approach preserves appealing metric space properties of Williams et al.'s~\cite{Williams2021} method.
Furthermore, we will see that alternative rotation-sensitive measures based on rectangular assignment algorithms~\cite{Crouse2016-ep} and semi-matching~\cite{Li2015_convergent} can produce unintuitive outcomes that our approach avoids.
Finally, we leverage our metric to show that the tuning of individual units is preserved above chance levels in deep layers of artificial and biological networks.
Thus, we present a quantitative approach to adjudicate between the competing hypotheses that ``tuning matters'' vs. ``geometry is all you need''~\cite{cosyne-workshop}, and we provide some emprical evidence in favor of the former hypothesis.

\section{Methods}

We use $\cO(N)$ and $\cP(N)$ to respectively denote the set of $N \times N$ orthogonal matrices and $N \times N$ permutation matrices.
These are commonly referred to as the \textit{orthogonal group} and \textit{permutation group}, respectively.
The permutation group is formally defined as follows:
\begin{equation}
\cP(N) = \Bigg \{ \mbP \in \reals^{N \times N} ~~ {\Bigg\vert} ~~ \begin{tabular}{ c c }
\small $\sum_{i} P_{ij} = 1$ & \small $\forall \, j \in \{ 1 \dots N \}$ \\[.35em] 
\small $\sum_{j} P_{ij} = 1$ & \small $\forall \, i \in \{ 1 \dots N \} $ \\[.35em] 
\small $P_{ij} \in \{0, 1\}$ & \small $\forall \, i, j \in \{ 1 \dots N \} \times \{ 1 \dots N \}$
\end{tabular}
\Bigg \}
\end{equation}
In words, a ``permutation matrix'' is a square matrix defined by containing only zeros and ones, and having each row and column sum to one.
For any $\mbP \in \cP(N)$ and $\mbv \in \reals^N$, the matrix-vector multiplication $\mbP\mbv$ outputs a vector with permuted elements.
It is easy to check that every permutation matrix is orthogonal, $\mbP^\top \mbP = \mbP \mbP^\top = \mbI$.
In other words, $\cP(N)$ is a subset of $\cO(N)$.

Essential to our approach is the relationship between permutation matrices and the set of \textit{doubly stochastic matrices}.
A doubly stochastic matrix is any square, nonnegative matrix whose rows and columns sum to one.
The set of doubly stochastic matrices forms a polytope, known as the \textit{Birkoff polytope}.
We denote the $N$-dimensional Birkhoff polytope as $\cB(N)$, formally:
\begin{equation}
\label{eq:birkhoff-polytope}
\cB(N) = \Bigg \{ \mbP \in \reals^{N \times N} ~~ {\Bigg\vert} ~~ \begin{tabular}{ c c }
\small $\sum_{i} P_{ij} = 1$ & \small $\forall \, j \in \{ 1 \dots N \}$ \\[.35em] 
\small $\sum_{j} P_{ij} = 1$ & \small $\forall \, i \in \{ 1 \dots N \} $ \\[.35em] 
\small $P_{ij} \geq 0$ & \small $\forall \, i, j \in \{ 1 \dots N \} \times \{ 1 \dots N \}$
\end{tabular}
\Bigg \}
\end{equation}
The Birkoff polytope is a convex set.
That is, for any two doubly stochastic matrices $\mbP_1 \in \cB(N)$ and $\mbP_2 \in \cB(N)$, we can define ${\mbP_3 = \alpha \mbP_1 + (1 - \alpha) \mbP_2}$ for arbitrary ${0 \leq \alpha \leq 1}$, and be guaranteed that $\mbP_3 \in \cB(N)$.
The celebrated \textit{Birkhoff–von Neumann theorem}~\cite{birkhoff1946three,vonNeumann} states that the vertices of $\cB(N)$ are one-to-one with $\cP(N)$, and this relationship will play an important role in our story.

Finally, we will be interested in generalizing the Birkhoff polytope to rectangular matrices, as this will help us generalize permutations to ``soft permutations.''
Specifically, consider a nonnegative matrix $\mbP \in \reals^{N_x \times N_y}$ whose rows each sum to $1 / N_x$ and whose columns each sum to $1 / N_y$.
Thus, the sum of all entries is equal to one.
The set of all such matrices defines a \textit{transportation polytope}~\cite{de2013combinatorics}; we denote this set as $\cT(N_x, N_y)$ and define it formally:
\begin{equation}
\label{eq:transportation-polytope}
\cT(N_x, N_y) = \Bigg \{ \mbP \in \reals^{N_x \times N_y} ~~ {\Bigg\vert} ~~ \begin{tabular}{ c c }
\small $\sum_{i} P_{ij} = 1 / N_y$ & \small $\forall \, j \in \{ 1 \dots N_y \}$ \\[.35em] 
\small $\sum_{j} P_{ij} = 1 / N_x$ & \small $\forall \, i \in \{ 1 \dots N_x \} $ \\[.35em] 
\small $P_{ij} \geq 0$ & \small $\forall \, i, j \in \{ 1 \dots N_x \} \times \{ 1 \dots N_y \}$
\end{tabular}
\Bigg \}
\end{equation}
Note that when $N = N_x = N_y$, all rows and columns sum to $1/N$. Thus, for any $\mbP \in \cB(N)$ we have that $(1/N)\mbP \in \cT(N, N)$.
In other words, except for a minor re-scaling factor, the Birkhoff polytope is a special case of the transportation polytope we defined.

\subsection{Problem Setup and Procrustes Distance}

To study the problem of comparing neural activations from different networks, we adopt a similar setting described in prior works (e.g. \cite{Raghu2017,Kornblith2019,Williams2021}). 
Specifically, neural population response vectors are sampled from two networks over a set of $M$ stimulus inputs.
We collect these responses into two matrices $\mbX \in \reals^{M \times N_x}$ and $\mbY \in \reals^{M \times N_y}$, where $N_x$ and $N_y$ denote the respective number of neurons in each network.
Our goal is to come up with methods to quantify the (dis)similiarity between $\mbX$ and $\mbY$ while ignoring nuisance transformations.
For example, CKA~\cite{Kornblith2019} and Procrustes shape distance~\cite{Williams2021} are invariant to translations, isotropic scalings, rotations, and reflections as nuisance transformations.
Other measures based on linear regression~\cite{Yamins2014}, partial least squares~\cite{Schrimpf2020}, and CCA~\cite{Raghu2017,Safaie2022} are invariant to a broader class of nuisance transformations: namely, invertible affine transformations within linear subspaces that account for a high fraction of (co)variance.

Consider the Procrustes shape distance as a concrete example.
For convenience, we assume that $\mbX$ and $\mbY$ have been pre-processed so that their columns sum to zero and $\Vert \mbX \Vert_F = \Vert \mbY \Vert_F = 1$.
Such pre-processing is necessary to remove the effects of translations and isotropic rescalings.
Assuming this pre-processing has been imposed and the two networks have the same number of neurons (i.e. ${N = N_x = N_y}$), then the Procrustes distance, which we denote $d_\cO$, can be defined as:
\begin{equation}
\label{eq:procrustes-1}
d_{\cO}(\mbX, \mbY) = \min_{\mbQ \in \cO(N)} \Vert \mbX - \mbY \mbQ \Vert_F
\end{equation}
Intuitively, the Procrustes distance is the minimal Euclidean distance between between $\mbX$ and $\mbY$ after optimizing over an orthogonal $N \times N$ alignment matrix $\mbQ$.
Thus, it is easy to see that the distance is invariant to rotations and reflections of the neural responses in $N$-dimensional space.
Due to this appealing geometric interpretation, eq.~(\ref{eq:procrustes-1}) is probably the most common definition of Procrustes distance.
However, this definition does not apply when $N_x \neq N_y$ since it would require us to subtract two matrices with different numbers of columns.
To remedy this, \textcite{Williams2021} suggested that $\mbX$ and $\mbY$ could be embedded into the same dimension by principal components analysis or zero-padding.
A more elegant solution is to simply define the Procrustes distance in the following manner, which is valid for $N_x \neq N_y$:
\begin{equation}
\label{eq:procrustes-2}
d_{\cO}(\mbX, \mbY) = \sqrt{\Tr[\mbX^\top \mbX] + \Tr[\mbY^\top \mbY] - 2 \Vert \mbX^\top \mbY \Vert_*}
\end{equation}
where $\Vert \mbM \Vert_*$ denotes the nuclear norm, or Schatten 1-norm, which is equal to the sum of the singular values of the matrix $\mbM$.
Importantly, \cref{eq:procrustes-1,eq:procrustes-2} are equivalent when $N_x = N_y$.
This fact is well-established, but we provide a self-contained proof in Supplement~\ref{proof:procrustes-1-2} for convenience.

An appealing property of the Procrustes distance is that it defines a \textit{metric space}.
More formally, given a class of nuisance transformations $\cF$, a distance function $d$ defines a metric space over equivalence classes associated to $\cF$ if it satisfies:
\begin{align}
\label{eq:metric-space-equivalence}
d \left(\mbX, \mbY \right) &= 0 \quad \text{if and only if there exists $f \in \cF$ such that}~\mbX = f(\mbY) \\
\label{eq:metric-space-symmetry}
d\left( \mbX, \mbY \right) &= d\left(\mbY, \mbX \right)\\
\label{eq:metric-space-triangle}
d\left(\mbX, \mbY \right) &\leq d\left(\mbX, \mbZ \right) + d\left(\mbZ, \mbY \right)
\end{align}
for any $\mbX \in \reals^{M \times N_x}, \mbY \in \reals^{M \times N_y}, \mbZ \in \reals^{M \times N_z}$.
It can be shown that Procrustes distance satisfies these properties, with $\cF$ corresponding to the set of translations, isotropic scalings, rotations, and reflections \cite{Williams2021}.

\subsection{One-to-One Matching Distance}

The purpose of this paper is to investigate alternative metrics that are not invariant to general orthogonal transformations (like Procrustes distance), but are still invariant to permutations of the neuron indices.
This is motivated by the hypothesis that neurons are usually arbitrarily indexed, but the tuning of individual units may be reproducible across networks.

\begin{figure}
\includegraphics[width=\linewidth]{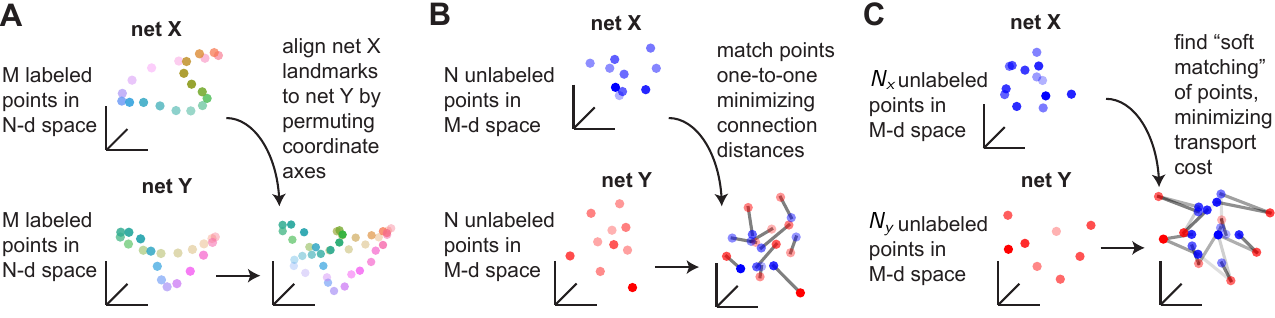} 
\centering
\caption{
    \textbf{(A)} One-to-one matching distance, schematized as alignment of $M$ points in $N$-dimensional space by optimally permuting coordinate axes. Colors denote landmark labels (e.g. image labels) that are common across the two networks.
    \textbf{(B)} Dual perspective of one-to-one matching distance, schematized as matching of $N$ unlabelled points in $M$-dimensional space. 
    \textbf{(C)} Soft matching distance generalizes the picture in panel B, and can be viewed as an optimal transport distance (see main text).
}
\label{fig:distance-schematic}
\end{figure}

When comparing networks with the same number of neurons ($N_x = N_y = N$), there is a natural generalization Procrustes distance which we call the \textit{one-to-one matching distance} (Fig.~\ref{fig:distance-schematic}A):
\begin{equation}
\label{eq:one-to-one-matching-distance-1}
d_{\cP}(\mbX, \mbY) = \min_{\mbP \in \cP(N)} \Vert \mbX - \mbY \mbP \Vert_F
\end{equation}
The only difference with Procrustes distance is that the minimization is performed over the group of $N$-dimensional permutation matrices, $\cP(N)$, instead of $N$-dimensional orthogonal matrices, $\cO(N)$.
Others have referred to this quantity as the ``permutation Procrustes'' problem~\cite{gower2004procrustes}.
But we prefer one-to-one matching distance to avoid confusion between the two.
Like the Procrustes distance, the one-to-one matching distance is symmetric and satisfies the triangle inequality (see, e.g., \cite{Williams2021}).

It is well-known, although not immediately obvious, that the optimal permutation can be efficiently found.
Indeed, a brute-force enumeration of all $N!$ permutation matrices is impossible for even moderately sized networks, but one can show (see Supplement~\ref{proof:one-to-one-1-2}) that \cref{eq:one-to-one-matching-distance-1} is equivalent to:
\begin{equation}
\label{eq:one-to-one-matching-distance-2}
d_{\cP}(\mbX, \mbY) = \sqrt{\min_{\mbP \in \cB(N)} \textstyle\sum_{i,j} \mbP_{ij} \Vert \mbx_i - \mby_j \Vert^2_2}
\end{equation}
where the minimization is performed subject to the constraint that $\mbP$ is in the Birkhoff polytope, $\cB(N)$, defined in \cref{eq:birkhoff-polytope}.
Above, $\Vert \mbx_i - \mby_j \Vert^2_2$ denotes the squared Euclidean distance between column $i$ of $\mbX$ and column $j$ of $\mbY$.
In other words, $\Vert \mbx_i - \mby_j \Vert^2_2$ is a measure of distance between a tuning curve $i$ from network $\mbX$ and tuning curve $j$ from network $\mbY$.

To build intuition for \cref{eq:one-to-one-matching-distance-2}, we offer the following interpretation as a ``transportation problem.''
Specifically, suppose we have $N$ ``sender warehouses'' at locations $\{\mbx_1, \dots, \mbx_N\}$, each with one unit of raw material.
We would like to transport all of our material to a set of ``receiver warehouses'' $\{\mby_1, \dots, \mby_N\}$ as cheaply as possible, where $\Vert \mbx_i - \mby_j \Vert^2_2$ quantifies the cost of transporting one unit of material from warehouse $i$ to warehouse $j$.
Any sender warehouse can split its supply amongst multiple receivers, so long as every sender releases all of its supply, $\sum_{j} \mbP_{ij} = 1$ for all $i$, and every receiver ends with exactly one unit of supply, $\sum_{i} \mbP_{ij} = 1$ for all $j$.
The Birkhoff polytope $\cB(N)$ represents the set of all valid transportation plans within these constraints.
Except in degenerate cases, there is a unique optimum and it is somewhat intuitive that the best strategy forgoes the option to split supplies, and instead finds a one-to-one matching of sender-receiver pairs.
That is, the solution will be found at a vertex of the Birkhoff polytope, and therefore be a permutation matrix due to the Birkhoff–von Neumann theorem.
Figure 2B provides a schematic illustration of this ``dual'' view of the one-to-one matching distance (compare with Fig. 2A).
\Cref{eq:one-to-one-matching-distance-1} can be efficiently computed by linear programming solvers as well as many specialized polynomial-time algorithms~\cite{Burkard2012} (see ~\ref{comp:complexity}). 

\subsection{Soft Matching Distance}

The one-to-one matching distance (eqns.~\ref{eq:one-to-one-matching-distance-1},\ref{eq:one-to-one-matching-distance-2}) is not applicable when $N_x \neq N_y$, which crucially limits its utility.
In analogy to how \cref{eq:procrustes-2} adapted the Procrustes distance to handle unequal network sizes, we now seek a similar generalization of the one-to-one matching distance.
A natural way to do this is to modify the constraints of the minimization in \cref{eq:one-to-one-matching-distance-2}, to obtain:
\begin{equation}
\label{eq:soft-matching-distance}
d_\cT (\mbX, \mbY) = \sqrt{\min_{\mbP \in \cT(N_x, N_y)} ~ \textstyle\sum_{ij} \mbP_{ij} \Vert \mbx_i - \mby_j \Vert^2}
\end{equation}
where $\cT(N_x, N_y)$ is the transportation polytope defined in \cref{eq:transportation-polytope}.
The idea is that the transportation and Birkhoff polytopes are essentially equivalent when $N = N_x = N_y$, except for a minor re-scaling factor.
In particular, it is easy to verify that $d_\cP(\mbX, \mbY) = \sqrt{N} \cdot d_\cT(\mbX, \mbY)$ when $N_x = N_y = N$.
That is, when comparing two networks of equal size, the soft matching distance is equal to the one-to-one matching distance except for a constant factor of $\sqrt{N}$.

\Cref{eq:soft-matching-distance} involves ``soft matching'' neuron labels in the sense that every row and every column of the optimal $\mbP$ may have more than one non-zero element.
We can again interpret \cref{eq:soft-matching-distance} as a transportation problem.
In this scenario, we have $N_x$ sender warehouses, each with $1/N_x$ units of material, and $N_y$ receiver warehouses, each requiring $1 / N_y$ units.
The set of candidate solutions (i.e. feasible transport plans) is given by $\cT(N_x, N_y)$.
When $N_x \neq N_y$, it is clearly necessary to do some amount of splitting/aggregating of material across multiple receivers/senders to satisfy the constraints of the problem.
Figure 2C illustrates this scenario (compare with Fig. 2B).

Readers who are familiar with optimal transport theory~\cite{villani2009optimal,santambrogio2015optimal,Peyre2019} will quickly realize that $d_\cT$ is simply the 2-Wasserstein distance between a uniform mixture of Dirac masses at ${\{\mbx_1, \dots, \mbx_{N_x}\}}$ and a uniform mixture of Dirac masses at ${\{\mby_1, \dots, \mby_{N_y}\}}$.
This allows us to immediately conclude the that soft matching distance is symmetric and satisfies the triangle inequality, which have been cited as advantageous properties~\cite{Williams2021}.
This connection to optimal transport also raises many interesting extensions, such as quantifying representational dissimilarity with entropy-regularized transport divergences~\cite{feydy19a}, but we leave these possibilities to future work.

\subsection{Soft Matching Correlation Score and Comparison with Semi-Matching}

Suppose that the columns of $\mbX$ and $\mbY$ have been mean-centered and normalized to unit length.
Then $\mbx_i^\top \mby_j$ is the Pearson correlation between neuron $i$ in network $\mbX$ and neuron $j$ in network $\mbY$.
In this setting, we can formulate a \textit{soft matching correlation score} between two networks:
\begin{equation}
\label{eq:soft-matching-correlation-score}
s_\cT(\mbX, \mbY) = \max_{\mbP \in \cT(N_x, N_y)} \sum_{i,j} \mbP_{ij} \mbx_i^\top \mby_j
\end{equation}
When $N_x = N_y$, this has an appealing interpretation: $s_\cT(\mbX, \mbY)$ equals the average correlation between neurons after optimal one-to-one matching.
Clearly, $s_\cT$ is closely related to the soft matching distance, $d_\cT$, defined in \cref{eq:soft-matching-distance}.
In fact, one can show (see Supplement~\ref{supplement:soft-matching-dist-corr}) that the matrix $\mbP$ which minimizes the distance in \cref{eq:soft-matching-distance} is the same matrix that maximizes the correlation in \cref{eq:soft-matching-correlation-score}.
Although some may prefer to use $d_\cT$ as a metric satisfying the triangle inequality, others may prefer to interpret $s_\cT$ since it is normalized between zero and one.

How does the soft matching correlation score compare with alternatives?
Inspired by analyses in \textcite{Li2015_convergent}, we consider a similarity score based on ``semi-matching'' assignments:
\begin{equation}
\label{eq:semi-matching-similarity-score}
s_\text{semi}(\mbX, \mbY) = \frac{1}{N_x} \sum_{i=1}^{N_x} \max_{j \in \{1, \dots, N_y\}} \mbx_i^\top \mby_j
\end{equation}
which is essentially the average correlation after matching every neuron in $\mbX$ to its most similar partner in $\mbY$.
Thus, each neuron in $\mbY$ may be matched to multiple neurons in $\mbX$ or matched to no partners at all.
Alternatively, assuming that $N_y \geq N_x$, one could define: 
\begin{equation}
\cR(N_x, N_y) = \Bigg \{ \mbP \in \reals^{N_x \times N_y} ~~ {\Bigg\vert} ~~ \begin{tabular}{ c c }
\small $\sum_{i} P_{ij} \leq 1$ & \small $\forall \, j \in \{ 1 \dots N_y \}$ \\[.35em] 
\small $\sum_{j} P_{ij} = 1$ & \small $\forall \, i \in \{ 1 \dots N_x \} $ \\[.35em] 
\small $P_{ij} \in \{0, 1\}$ & \small $\forall \, i, j \in \{ 1 \dots N_x \} \times \{ 1 \dots N_y \}$
\end{tabular}
\Bigg \}
\end{equation}
as the set of allowable matchings.
Here, every neuron in $\mbX$ is matched to one neuron in $\mbY$, and each neuron in $\mbY$ is matched to one or zero neurons in $\mbX$.
Then, we can define:
\begin{equation}
\label{eq:rectangular-matching-correlation-score}
s_\cR(\mbX, \mbY) = \max_{\mbP \in \cR(N_x, N_y)} \frac{1}{N_x} \sum_{i,j} \mbP_{ij} \mbx_i^\top \mby_j
\end{equation}
as a similarity score.
Again, this only applies when $N_y \geq N_x$, since there are no feasible matchings when $N_x > N_y$.
\textcite{Crouse2016} describes algorithms for solving the maximization in \cref{eq:rectangular-matching-correlation-score}.

\begin{figure}
\includegraphics[width=\linewidth]{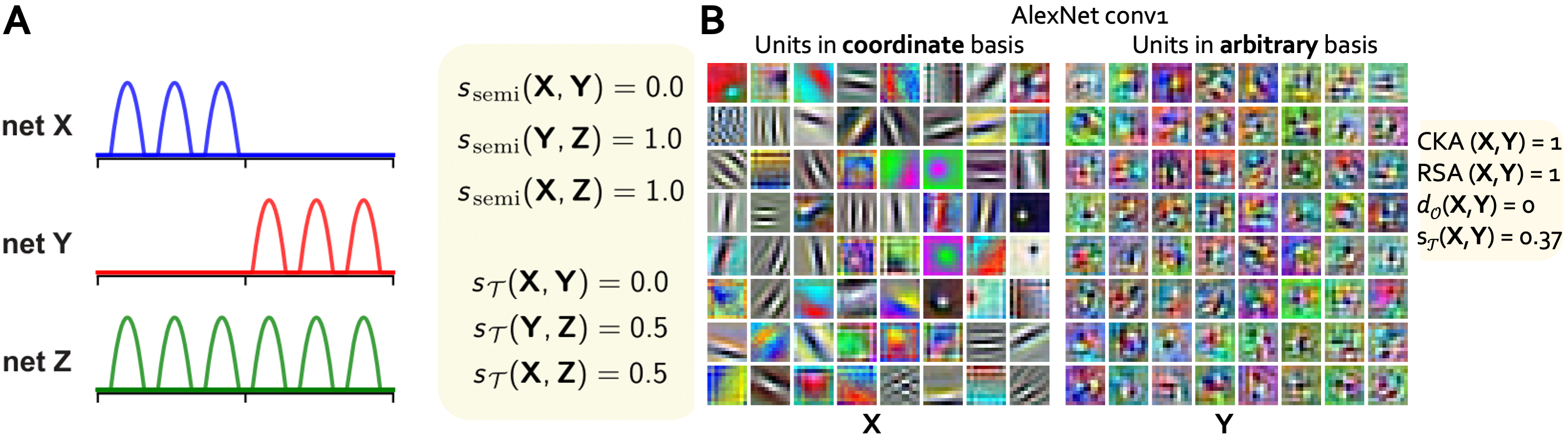} 
\centering
\caption{
    \textbf{(A)} (left) Three networks ($\mbX$, $\mbY$, and $\mbZ$), composed of 1D tuning curves (network sizes: 3, 3, and 6).
    (right) Similarity scores for networks in panel A quantified by $s_\text{semi}$ and $s_\cT$. In this example, each tuning curve is normalized to unit length but not mean-centered.
    \textbf{(B)} Visualization of the 64 {\tt conv1} filters of size $11\times 11 \times 3$ in AlexNet trained on ImageNet in (left) standard and (right) rotated basis. The rotation matrix is sampled uniformly from the SO(64) group. 
}
\label{fig:3}
\end{figure}

It is easy to see that of the three similarity scores, $s_\cT$, $s_\text{semi}$, and $s_\cR$, only the soft matching score is symmetric $s_\cT(\mbX, \mbY) = s_\cT(\mbY, \mbX)$.
Moreover, both $s_\text{semi}$ and $s_\cR$ will tend to view a network with many units (e.g. a ``wide'' deep net layer) as similar to everything it is compared with.
This is illustrated in Figure~\ref{fig:3}A, which gives a simple example where $\mbX$ and $\mbY$ are decorrelated, $s_{\text{semi}}(\mbX, \mbY) = 0$, but at the same time $\mbX$ and $\mbY$ are both maximally correlated to a third network, ${s_{\text{semi}}(\mbX, \mbZ) = s_{\text{semi}}(\mbY, \mbZ) = 1}$.
Put differently, on the basis of the $s_\text{semi}$ similarity scores, it is tempting to observe \textit{``$\mbX$ is perfectly correlated to $\mbZ$, and $\mbY$ is perfectly correlated to $\mbZ$''} and then falsely conclude that \textit{``$\mbX$ is perfectly correlated to $\mbY$.''}
This counter-intuitive behavior also applies to $s_\cR$, which behaves identically to $s_\text{semi}$ in this example.
In contrast, the soft matching correlation score gives a more intuitive result: it treats $\mbZ$ as only 50\% correlated to $\mbX$ and $\mbY$ (Fig.~\ref{fig:3}A).

\section{Applications}

\subsection{Highlighting the Limitations of Existing (Dis)similarity Metrics for Single-Neuron Tuning}
We first illustrate how prevailing (dis)similarity measures fall short in capturing essential properties of single-neuron tuning due to their rotational invariance. To exemplify this point, we visualize filter weights from the first convolutional layer ({\tt conv1}) of AlexNet trained on ImageNet. Prior research has consistently demonstrated the emergence of Gabor (or ``edge-detecting") filters in the initial layers of deep convolutional networks trained on natural image datasets~\cite{krizhevsky2012imagenet, olah2020overview}, a phenomenon clearly visible in our analysis (Fig.~\ref{fig:3}B, left). Intuitively, the structure within these individual filters specifies a coordinate basis in neural activation space that is special and non-random. Indeed, when we apply a random rotation to examine the same weights in an arbitrary basis ($\mbY$) (Fig.~\ref{fig:3}B, right), the tuning of individual units differs significantly from that in the coordinate basis ($\mbX$). In particular, we see minimal signatures of edge-detecting filters in $\mbY$.

It should be emphasized that Fig.~\ref{fig:3}B visualizes filter weights and applies a random rotation in weight space.
Since it is difficult to interpret weights in deeper layers, measures of representational (dis)similarity are usually computed on neural activations instead of weights.
However, for the first layer of the network the weights and activations are very closely related: rotating the filter weights (as done in Fig.~\ref{fig:3}B, \textit{right}) is equivalent to rotating the neural pre-nonlinearity neural activations.
Thus, measures like Procrustes distance, CKA, and CCA all fail to capture the distinction between $\mbX$ and $\mbY$, treating them as equivalent representations. In contrast, the soft-matching distance effectively distinguishes them. Given the sensitivity of the soft-matching distance to single-neuron tuning, beyond mere similarities in representational geometry, we propose that it may also prove valuable in the quantification of disentangled representation learning (see~\ref{drl}). 

\subsection{Evidence for privileged coordinate bases in deep hidden layer representations}

Figure~\ref{fig:3} presents evidence that individual neural tuning functions are non-arbitrary in the first layer of a deep network---a finding that has been documented in past work~\cite{krizhevsky2012imagenet, olah2020overview}.
The soft-matching distance enables us to precisely quantify this effect and investigate the extent to which the coordinate axes (i.e. individual neural tuning curves) are reproducible in deeper hidden layers.
That is, we hypothesize that neural networks trained from different initial weights will converge onto similar tuning curves.
This \textit{convergent basis hypothesis} (``tuning matters'') can be contrasted with the \textit{arbitrary basis hypothesis} (``geometry is all you need''), which predicts that two networks may converge onto similar representational subspaces but with arbitrarily rotated coordinate axes.
As already mentioned, this latter hypothesis is conceptually aligned with existing (dis)similarity measures that are rotation-invariant (see e.g. \cite{Kornblith2019}).

To adjudicate between these hypotheses, we employed the soft-matching similarity metric to explore how the alignment between different neural representations evolves with a gradual change of basis. We compare early and late layer representations in deep convolutional networks trained on object categorization with different random initializations, different architectures (ResNet20/VGG16)~\cite{he2016deep, simonyan2014very} and on different datasets (CIFAR10/CIFAR100)~\cite{krizhevsky2009learning} to explore whether the bases in different networks share a non-random relationship with respect to each other (evidence for the \textit{convergent basis hypothesis}) or whether this relationship is arbitrary (evidence for the \textit{arbitrary basis hypothesis}). We emphasize that such an inquiry necessitates the development of rotation-sensitive measures, as measures invariant to rotations would exhibit no change with a basis shift. 

\begin{figure}
\includegraphics[width=\linewidth]{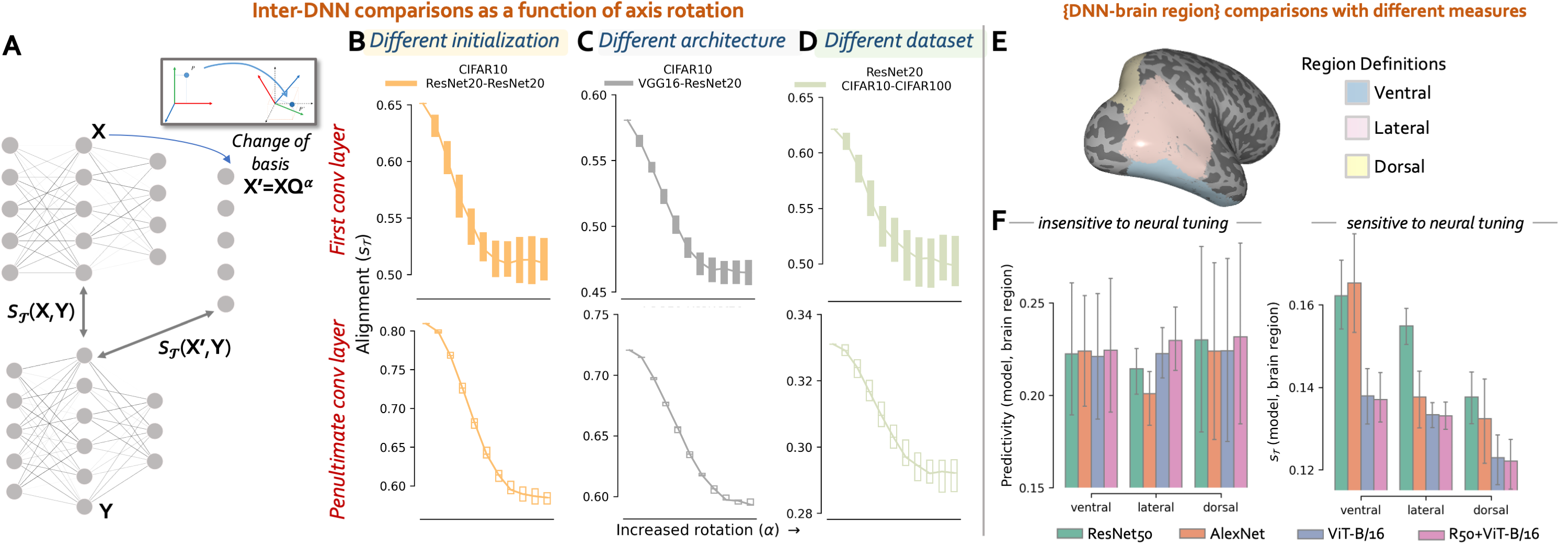} 
\centering
\caption{
    \textbf{Soft-matching similarity reveals the non-arbitrariness of activation bases in DNNs.} We investigate how changes in activation bases impact the soft-matching similarity between two neural representations, $\mbX$ and $\mbY$ (A). Panels B, C and D depict the alignment between networks trained with different initial random seeds (with a fixed ResNet20 architecture), different architectures (ResNet20 vs. VGG16) and different datasets (CIFAR10 vs. CIFAR100) as a function of axis rotation, respectively.
    \textbf{Comparison of DNNs trained on image recognition with the three high-level visual stream representations using different (dis)similarity measures} (E) Definition of the visual streams on a cortical map. (F) DNN-brain region alignment measured using (left) linear predictivity and the soft-matching similarity score ($s_\cT$).
    }  
\label{fig:5}
\end{figure}

Our approach is as follows: We sample a random rotation matrix $\mbQ$ uniformly over the special orthogonal group $SO(N)$~\cite{Ozols2009HowTG}. We then instantiate a sequence of rotation matrices that interpolate between $\mbQ$ and the identity matrix ($\mbI$) along the $SO(N)$ manifold. To achieve this, we calculate fractional powers $\mbQ^\alpha = \exp[\alpha \cdot \log [\mbQ]]$, where $0 \leq \alpha \leq 1$ and $\exp[\cdot]$ and $\log[\cdot]$ denote the matrix exponential and matrix logarithm, respectively. Intermediate values of $\alpha$ smoothly interpolate between that $\mbQ$ ($\alpha = 1$) and $\mbI$ ($\alpha = 0$).
Thus, varying $\alpha$ smoothly varies the degree of rotation, interpolating between a random basis and the original basis of the neural representation.
For each pair of representations ($\mbX, \mbY$), we alter the basis of one representation, say $\mbX$, by multiplying on the right by $\mbQ^\alpha$. We then measure how the soft-matching correlation score, ${s_\cT(\mbX \mbQ^\alpha, \mbY)}$, changes as the degree of rotation is smoothly increased from $\alpha=0$ to $\alpha=1$ (Fig.~\ref{fig:5}A). If the similarity decreases monotonically as a function of $\alpha$, it furnishes empirical evidence supporting the convergent basis hypothesis.

Our empirical findings are illustrated in Fig.~\ref{fig:5}B-D. In panel B, we observe that representations from the same network (specifically, ResNet20 trained on CIFAR10) but with varying initial random seeds are significantly more aligned in the standard (coordinate) basis as compared to an arbitrary basis, and this alignment decreases gradually with increasing rotation ($\alpha$). This trend holds for both early (top) and late (bottom) convolutional layers. This provides evidence that deep convolutional networks trained on image data from different initial weights tend to converge onto similar (though obviously not identical) bases. Strikingly, this trend holds even when comparing networks trained with different architectures (ResNet20 vs. VGG16, Fig.~\ref{fig:5}C) as well as networks trained on different datasets (CIFAR10 vs. CIFAR100, Fig.~\ref{fig:5}D). While the strict, one-to-one matching distance (eq.~\ref{eq:one-to-one-matching-distance-1}) can be used to quantify these trends across networks with the same architecture, the soft matching distance (eq.~\ref{eq:soft-matching-distance}) or soft matching correlation (eq.~\ref{eq:soft-matching-correlation-score}) developed in this paper are needed to rigorously quantify this effect across different architectures (as in Fig.~\ref{fig:5}C). Overall, our results suggest that the coordinate axes of neural representations across diverse network architectures and training diets are aligned at above-chance levels, lending support for the \textit{convergent basis hypothesis}. 

\subsection{Soft-matching distance meaningfully distinguishes between neuroscientific hypotheses}
Beyond comparing similarity of representations across different artificial neural networks, there has been a recent surge of interest in applying (dis)similarity measures to compare artificial and biological neural networks~\cite{barrett2019analyzing, Schrimpf2020}. We next demonstrate an application of this metric for model-brain comparisons in neuroscience. In particular, numerous studies over the last decade have revealed that deep networks optimized for behaviorally relevant goals like object categorization learn internal representations that are similar to those in the macaque inferotemporal cortex (IT) (or the ventral visual stream in humans)~\cite{Yamins2014, khaligh2014deep}, an area believed to support object recognition in primates~\cite{dicarlo2012does}. Such findings have sparked optimism that by comparing deep network and neurobiological activation statistics, we can explain specific characteristics of the brain as optimized solutions for specific computational problems faced by organisms~\cite{yamins2016using, kanwisher2023using}. 

However, recent studies have revealed several counter-intuitive results. For instance, deep networks trained on object categorization were shown to not only accurately model representations in the ventral visual pathway but also in the dorsal and lateral visual pathways~\cite{finzi2022deep}. Traditionally, the latter visual streams were hypothesized to be engaged in distinct functions, such as action recognition, social perception, or visually-guided action~\cite{pitcher2021evidence, goodale1992separate}. Yet another counter-intuitive finding is that vision transformers~\cite{dosovitskiy2020image} exhibit equivalent performance to convolutional networks in predicting biological responses ~\cite{conwell2022can}, even though the latter were designed with some inspiration from biological systems. All of this raises a critical question: are deep networks optimized for object categorization, irrespective of the biological plausibility of their architecture, equally viable models for all three of these visual streams? Or do these findings indicate an inadequacy in our current tools and metrics when it comes to distinguishing between these neuroscientific hypotheses? 

One potential explanation is that existing representational (dis)similarity measures may be too permissive to differentiate between various hypotheses. More stringent measures which exclusively seek invariance only with respect to neuron permutations might unveil a different finding. To test this hypothesis, we leverage the massive Natural Scenes Dataset (NSD) and compare model and fMRI responses across the three visual streams using both a measure that maintains invariance under all invertible linear transformations (linear predictivity, $R^2$), and our proposed soft-matching distance. We conduct our comparative analyses using the shared set of 1,000 images, each viewed three times by four different participants.  We extract feature representations from four candidate neural architectures, all of which were trained for object categorization on ImageNet. This set comprises two DCNNs, namely ResNet50 and AlexNet, as well as two vision transformers~\cite{dosovitskiy2020image}, ViT-B/16 and a hybrid model (R50+ViT-B/16). In the hybrid model, the input sequence to the ViT is formed from intermediate feature maps of a regular ResNet50. We compare the penultimate representation from each model to the measured brain activity in each of the three high-level visual streams using the soft-matching metrics and linear predictivity. The latter is computed by fitting an $l_2$ regularized linear regression model on the model representations to predict the measured brain activity using a 70/10/20 train/validation/test split. The predictivity is quantified as the Pearson's correlation coefficient (R) between the measured and predicted responses of each brain voxel, averaged within each stream. The regularization parameter was optimized independently for each subject and each high-level visual stream by testing among 8 log-spaced values in \texttt{[1e-4, 1e4]}.   

Comparing representations across model and brain region combinations, we observed that linear predictivity ($R^2$) was insufficient to discern differences between different architectures (CNNs or transformers) and distinct visual processing streams: all model-region scores exhibited  similar ranking (Fig.~\ref{fig:5}F). On the other hand, the soft-matching metrics proved effective in adjudicating between models and revealed significant differences (i)  among CNNs vs. transformer architectures in terms of their similarity to brain representations, and (ii) in the ability of object categorization models to capture representations within the three putative visual streams. According to this metric, convolutional networks emerged as superior models for modeling the ventral visual stream when compared to transformers. Furthermore, object categorization models, regardless of their architectural design, demonstrated a better fit with the ventral visual stream as opposed to the other streams. These conclusions are in line with our intuitive understanding of these models and neural systems.


\section{Conclusion}

We leveraged  concepts from optimal transport theory to introduce a metric on neural representations that is rotation-sensitive but permutation-invariant. This metric, which we call \textit{soft matching distance}, generalizes the one-to-one matching distance proposed in~\cite{Williams2021} to networks of varying sizes by employing \textit{soft permutation} or \textit{soft assignment} of neurons~\cite{rangarajan1997softassign,singh2020model,li2020representation,grave2019unsupervised,Alvarez2019}.

The soft matching metric reveals structure that is invisible to popular rotation-insensitive measures like CKA~\cite{Kornblith2019}, RSA~\cite{Kriegeskorte2008}, and Procrustes distance~\cite{Williams2021}. For example, our experiments on CIFAR10 and CIFAR100 leverage soft matching distance to show reproducible convergence of activation bases across networks. This convergence holds true across various factors such as initial random seeds, architectural differences, and training diets. These findings are consistent with a variety of anecdotal accounts within the interpretable deep learning literature, such as ``curve detector'' units~\cite{cammarata2020curve}, shape tuning~\cite{Pospisil2018}, and object detectors~\cite{zhou2014object}.
Despite these examples, most hidden layer units are not (as far as we can tell) semantically meaningful---yet, these non-semantic units are still important for network performance~\cite{Morcos2018}.
The soft matching distance therefore addresses an important need to quantitatively compare single unit tuning across networks without reliance on semantic labels.

Moreover, we extend the utility of our metric to the domain of comparing artificial and biological neural networks. We observed that our metric outperforms measures with more inherent invariances, such as linear predictivity, in terms of distinguishing between models. This development offers neuroscientists an additional---and potentially more discerning---tool to interrogate the commonalities and distinctions between biological and artificial networks. 

Why might networks have a distinguished and convergent basis, and why is the basis often overlooked in practice? Prior studies justify the use of rotation-invariant metrics since a network layer is arbitrary up to a full-rank matrix multiplication: hence, the subsequent weight matrix can absorb the matrix inverse, rendering the choice of basis immaterial~\cite{Kornblith2019}. However, this disregards the fact that nonlinearities, such as Rectified Linear Units (ReLUs), are applied along particular dimensions (i.e. the coordinate axes) within the space of neural activations. These non-linearities might thus act as symmetry-breaking mechanisms that favor certain activation bases over others.

Overall, our work develops a metric, the \textit{soft matching distance}, which complements existing measures of representational similarity like CKA, RSA, and Procrustes shape distance.
Relative to these existing rotation-invariant measures, this new distance is better suited to interrogate representations at the level of single-neuron tuning.
Our applications of the metric thus far support the view that single neuron tuning is preserved above chance levels across networks, which may be an important clue into the complex computations performed within artificial and biological neural circuits.

\section*{Acknowledgements}

We thank Nancy Kanwisher and David Alvarez-Melis for insightful discussions and comments on the manuscript.

\printbibliography

\clearpage
\setcounter{page}{1}
\begin{appendices}


\vspace{-2em}
\section{Supplementary Results and Comments}

\subsection{Proof that \cref{eq:procrustes-1,eq:procrustes-2} are equivalent when $N_x = N_y = N$}
\label{proof:procrustes-1-2}

This is the result of a straightforward calculation, exploiting several elementary facts from linear algebra.
First, for any matrix $\mbA$ we have that $\Vert \mbA \Vert_F^2 = \Tr[ \mbA^\top \mbA ]$.
Second, for any matrices $\{\mbA, \mbB, \mbC\}$ with appropriate dimensions such that the product $\mbA \mbB \mbC$ is defined, we have that $\Tr[\mbA \mbB \mbC] = \Tr[\mbC \mbA \mbB] = \Tr[\mbB \mbC \mbA]$, which is called the \textit{cyclic trace property}.
Finally, for any orthogonal matrix $\mbQ \in \cO(N)$ we have $\mbQ^\top \mbQ = \mbQ \mbQ^\top = \mbI$.

With these ingredients we can manipulate the squared Procrustes distance as follows:
\begin{align*}
d_\cO^2(\mbX, \mbY) &= \min_{\mbQ \in \cO(N)} \Vert \mbX - \mbY \mbQ \Vert_F^2 \\
&= \min_{\mbQ \in \cO(N)} \Tr[\mbX^\top \mbX + \mbQ^\top \mbY^\top \mbY \mbQ - 2 \mbX^\top \mbY \mbQ] && \text{($\Vert \mbA \Vert_F^2 = \Tr[ \mbA^\top \mbA ]$)} \\
&= \min_{\mbQ \in \cO(N)} \Tr[\mbX^\top \mbX] + \Tr [ \mbQ^\top \mbY^\top \mbY \mbQ ] - 2 \Tr [ \mbX^\top \mbY \mbQ] && \text{($\Tr[\cdot]$ is linear)} \\
&= \min_{\mbQ \in \cO(N)} \Tr[\mbX^\top \mbX] + \Tr [ \mbY^\top \mbY \mbQ \mbQ^\top ] - 2 \Tr [ \mbX^\top \mbY \mbQ] && \text{(cyclic trace property)}\\
&= \min_{\mbQ \in \cO(N)} \Tr[\mbX^\top \mbX] + \Tr [ \mbY^\top \mbY ] - 2 \Tr [ \mbX^\top \mbY \mbQ] && \text{(orthogonality)}\\
&= \Tr[\mbX^\top \mbX] + \Tr [ \mbY^\top \mbY ] - 2 \max_{\mbQ \in \cO(N)} \Tr [ \mbX^\top \mbY \mbQ] && \text{(first two terms are constant)}\\
&= \Tr[\mbX^\top \mbX] + \Tr [ \mbY^\top \mbY ] - 2 \Vert \mbX^\top \mbY \Vert_* && \text{(discussed below)}
\end{align*}
as claimed in the main text.
The final step is the comes from the celebrated closed form solution to the orthogonal Procrustes problem, which is comprehensively reviewed by \textcite{gower2004procrustes}.
Briefly, the result can be understood by considering the singular value decomposition $\mbX^\top \mbY = \mbU \mbS \mbV^\top$.
Then, due to the cyclic trace property,
\begin{equation}
\Tr [ \mbX^\top \mbY \mbQ] = \Tr [ \mbU \mbS \mbV^\top \mbQ] = \Tr [ \mbS \mbV^\top \mbQ \mbU ]
\end{equation}
This final expression is maximized by setting $\mbQ = \mbV \mbU^\top$, resulting in:
\begin{equation}
\Tr [ \mbS \mbV^\top \mbV \mbU^\top \mbU ] = \Tr [ \mbS ] = \Vert \mbX^\top \mbY \Vert_*
\end{equation}
Since the sum of the diagonal elements of $\mbS$ is simply the sum of the singular values of $\mbX^\top \mbY$ (i.e. equal to the nuclear norm of this matrix).

\subsection{Proof that \cref{eq:one-to-one-matching-distance-1,eq:one-to-one-matching-distance-2} are equivalent}
\label{proof:one-to-one-1-2}

Recall that we are in the setting where $N_x = N_y = N$.
First, for any matrix $\mbA \in \reals^{M \times N}$ with columns $\{ \mba_1, \dots, \mba_N \}$ we have $\Vert \mbA \Vert_F^2 = \sum_{i=1}^N \Vert \mba_i \Vert^2$.
Since $\sum_j \mbP_{ij} \mby_j$ gives column $i$ of the matrix product $\mbY \mbP$, we have:
\begin{equation}
\Vert \mbX - \mbY \mbP \Vert_F^2 = \sum_{i=1}^N \Vert \mbx_i  - \sum_{j=1}^N \mbP_{ij} \mby_j \Vert^2
\end{equation}
Recall that $\mbP$ is a permutation matrix in the present context.
Thus, let $\sigma(i) \in \{1, \dots, N\}$ denote the index of the unique nonzero element of row $i$ in $\mbP$.
Intuitively, $\sigma(i)$ defines the permutation in which column $i$ of $\mbX$ is matched to column $\sigma(i)$ of $\mbY$.
With this notation, we can re-write the expression above:
\begin{equation}
\sum_{i=1}^N \Vert \mbx_i  - \sum_{j=1}^N \mbP_{ij} \mby_j \Vert^2 = \sum_{i=1}^N \Vert \mbx_i  - \mby_{\sigma(i)} \Vert^2
\end{equation}
Now define $\delta[i, j]$ as a function that takes in two integers and evaluates to one if $i = j$ and evaluates to zero if $i \neq j$.
(This is often called the Kronecker delta function.)
We can write:
\begin{equation}
\sum_{i=1}^N \Vert \mbx_i  - \mby_{\sigma(i)} \Vert^2 = \sum_{i=1}^N \sum_{j=1}^N \delta[\sigma(i), j] \cdot \Vert \mbx_i  - \mby_j \Vert^2
\end{equation}
since the inner sum will evaluate to zero whenever $j \neq \sigma(i)$.
Finally, we argue that $\mbP_{ij} = \delta[\sigma(i), j]$.
Indeed, $\sum_j \delta[\sigma(i), j] \mby_j = \mby_\sigma(i)$ which agrees with $\sum_j \mbP_{ij} \mby_j = \mby_\sigma(i)$.
Thus, we have shown that:
\begin{equation}
d_{\cP}^2(\mbX, \mbY) = \min_{\mbP \in \cP(N)} \Vert \mbX - \mbY \mbP \Vert_F^2 = \min_{\mbP \in \cP(N)} \sum_{i,j} \mbP_{ij} \Vert \mbx_i - \mby_j \Vert^2_2
\end{equation}
To arrive at \cref{eq:one-to-one-matching-distance-2}, we need to show that we can relax the constraint of the minimization over the permutation group to over the Birkhoff polytope---i.e. to show that:
\begin{equation}
\label{eq:supplement-equivalence-of-birkhoff-perm}
\min_{\mbP \in \cP(N)} \sum_{i,j} \mbP_{ij} \Vert \mbx_i - \mby_j \Vert^2_2 = \min_{\mbP \in \cB(N)} \sum_{i,j} \mbP_{ij} \Vert \mbx_i - \mby_j \Vert^2_2
\end{equation}
Here we evoke two well-known results.
First, the celebrated Birkhoff–von Neumann theorem states that the vertices of $\cB(N)$ are one-to-one matched with the permutation matrices $\cP(N)$.
Second, the final expression is a linear program since the objective function is linear in $\mbP$ and the constraints are linear (as can be verified from eq.~\ref{eq:birkhoff-polytope}).
Thus, we evoke a basic fact from the theory of linear programming (see e.g.~\cite{bertsimas1997introduction}), which states that, assuming that a finite solution exists, at least one vertex of the feasible set is a solution.
Any such vertex is called a ``basic feasible solution'' and the fact that such solutions exist motivates the well known simplex algorithm for linear programming.
Thus, we conclude that relaxing the constraints from $\mbP \in \cP(N)$ to $\mbP \in \cB(N)$ does not allow us to further minimize the objective, and so \cref{eq:supplement-equivalence-of-birkhoff-perm} is valid.
Taking square roots on both sides of \cref{eq:supplement-equivalence-of-birkhoff-perm} proves the result claimed in the main text.

\subsection{Relation between soft matching distance and correlation score}
\label{supplement:soft-matching-dist-corr}

Here we show that the optimal soft permutation matrix $\mbP \in \cT(N_x, N_y)$ that minimizes the expression in eq.~\ref{eq:soft-matching-distance} equals the one which maximizes the expression in eq.~\ref{eq:soft-matching-correlation-score}.
First, beginning with the minimization problem in eq.~\ref{eq:soft-matching-distance}, we can break the expression into three terms:
\begin{align}
&\argmin_{\mbP \in \cT(N_x, N_y)} \sum_{ij} \mbP_{ij} \Vert \mbx_i - \mby_j \Vert^2\\
&\hspace{1em}= \argmin_{\mbP \in \cT(N_x, N_y)} ~ \sum_{ij} \mbP_{ij} (\mbx_i^\top \mbx_i + \mby_j^\top \mby_j - 2 \mbx_i^\top \mby_j ) \\
&\hspace{1em}= \argmin_{\mbP \in \cT(N_x, N_y)} ~ \underbrace{\sum_{ij} \mbP_{ij} \mbx_i^\top \mbx_i}_{\text{(A)}} + \underbrace{\sum_{ij} \mbP_{ij} \mby_j^\top \mby_j}_{\text{(B)}} - 2 \underbrace{\sum_{ij} \mbP_{ij} \mbx_i^\top \mby_j}_{\text{(C)}}
\end{align}
Considering term (A) first, we argue that this term is constant with respect to any feasible $\mbP$ since:
\begin{equation}
\sum_{ij} \mbP_{ij} \mbx_i^\top \mbx_i = \sum_{i=1}^{N_x} \left ( \sum_{j=1}^{N_y} \mbP_{ij} \right ) \mbx_i^\top \mbx_i = \sum_{i=1}^{N_x}  \left ( \frac{1}{N_x} \right ) \mbx_i^\top \mbx_i
\end{equation}
where the final equality follows from definition of the transportation polytope in \cref{eq:transportation-polytope}---namely, the rows of $\mbP$ each sum to $1 / N_x$.
We then can make a similar argument for term (B).
In particular, since the columns of $\mbP$ each sum to $1 / N_y$, we have:
\begin{equation}
\sum_{ij} \mbP_{ij} \mby_j^\top \mby_j = \sum_{j=1}^{N_y} \left ( \sum_{i=1}^{N_x} \mbP_{ij} \right ) \mby_j^\top \mby_j = \sum_{j=1}^{N_y} \left ( \frac{1}{N_y} \right ) \mby_j^\top \mby_j
\end{equation}
In summary, we see that only term (C) is non-constant.
That is, we have
\begin{align}
\argmin_{\mbP \in \cT(N_x, N_y)} \sum_{ij} \mbP_{ij} \Vert \mbx_i - \mby_j \Vert^2 &= \argmin_{\mbP \in \cT(N_x, N_y)} -2 \sum_{ij} \mbP_{ij} \mbx_i^\top \mby_j + const. \\
&= \argmax_{\mbP \in \cT(N_x, N_y)} \sum_{ij} \mbP_{ij} \mbx_i^\top \mby_j
\end{align}
as we have claimed.

\subsection{Computational complexity} \label{comp:complexity}
Computing the soft-matching distance requires solving an optimal transport problem in the discrete setting. The solution to the transportation problem, which is a linear program, can be derived using the network simplex algorithm. With efficient implementations of the simplex algorithm as in the Python Optimal Transport Library, the complexity of solving the linear program is O($n^3 \log n$), assuming that the two representations being compared have $n$ units. Such efficient implementations enable broad application of optimal transport-based solutions in real-world settings. 

\subsection{Relevance of the Soft-Matching Metric to Disentangled Representation Learning Metrics}
\label{drl}
It is worth noting that the proposed soft-matching metric can also serve as a valuable tool in the field of disentangled representation learning (DRL)~\cite{bengio2013representation} due to its sensitivity to the representational basis. In DRL, the objective is to learn a model that effectively disentangles and makes the underlying generative factors of the data explicit in representational form (i.e. aligned with representational units). Within the DRL literature, various measures have been developed to quantify the alignment between learned representations and ground truth generative factors. Typically, the desiderata involve a combination of different criteria, encompassing the similarity in information content (explicitness) and the degree of one-to-one correspondence between the representational units and generative factors (modularity and compactness)~\cite{eastwood2018framework}. The soft-matching distance metric offers a unique advantage by simultaneously capturing sensitivity to both of these critical properties.

\end{appendices}

\end{document}